# Ideal Partition of Resources for Metareasoning*

**Eric J. Horvitz**

Medical Computer Science Group

Knowledge Systems Laboratory

Stanford University

Stanford, California 94305

**John S. Breese**

Rockwell International Science Center

Palo Alto Laboratory

444 High Street

Palo Alto, CA 94301

January 1990

## Abstract

We can achieve significant gains in the value of computation by metareasoning about the nature or extent of base-level problem solving before executing a solution. However, resources that are irrevocably committed to metareasoning are not available for executing a solution. Thus, it is important to determine the portion of resources we wish to apply to metareasoning and control versus to the execution of a solution plan. Recent research on rational agency has highlighted the importance of limiting the consumption of resources by metareasoning machinery. We shall introduce the *metareasoning-partition* problem—the problem of ideally apportioning costly reasoning resources to planning a solution versus applying resource to executing a solution to a problem. We exercise prototypical metareasoning-partition models to probe the relationships between time allocated to metareasoning and to execution for different problem classes. Finally, we examine the value of metareasoning in the context of our functional analyses.

---

*This work was supported by a NASA Fellowship under Grant NCC-220-51, by the National Science Foundation under Grant IRI-8703710, and by the U.S. Army Research Office under Grant P-25514-EL. Computing facilities were provided by the SUMEX-AIM Resource under NLM Grant LM05208.

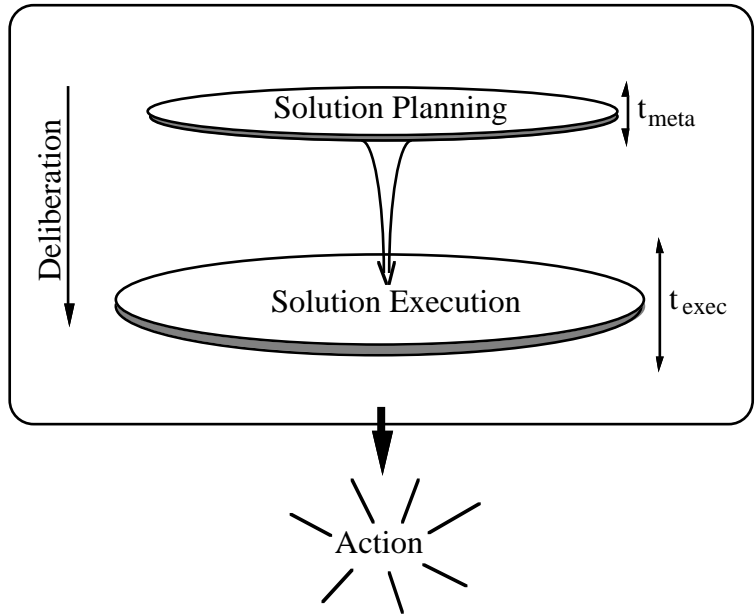

Figure 1: We are interested in determining the ideal quantity of resources that should be applied to metareasoning, given knowledge about the ability of metareasoning to reduce the difficulty of solving the base problem.

# 1 Introduction

Recent research, spanning AI and the decision sciences, has highlighted the usefulness of metareasoning for directing and limiting base-level problem solving [1, 9, 11]. Metareasoning refers to several classes of explicit consideration and control of problem-solving processes; these include planning a solution, limiting the extent of base-level problem solving, choosing the next best problem-solving step, and searching for a better formulation of a problem instance. Metareasoning can increase the efficiency or increase the utility of base-level inference. However metalevel deliberation can consume a significant portion of valuable computational resources.

Recent investigation of metareasoning and rational action has highlighted the importance of managing the consumption of resources by metareasoning processes. Although some of our earlier work relied on tractable solutions to metalevel decision problems [9], we have stressed the need to generalize analyses and implementations to the case where a metareasoner may dynamically choose to consume greater portions of the total quantity of resources used in problem solving [7]. Indeed, given a problem, a context, and the details of a computational architecture, it may be best to expend a significant proportion of the total consumed resources, at the metalevel. At other times, however, it may be best to expend little or no resources on metareasoning. We shall discuss the problem of ideally apportioning costly reasoning resources to metalevel inference versus applying resource to executing a solution to a problem. We call this the *metareasoning-partition* problem. This metareasoning-partition problem is an important link in the pursuit of rational behavior under resource constraints.

We shall develop insights about the ideal partition of resources for metareasoning by constructing and solving expressive mathematical models of the metareasoning-partition problem. These models capture prototypical functional relationships between reasoning and metareasoning for a large class of problems. We exercise the models to identify the ideal configuration of a small number of variables that define an agent's metareasoning policy. In a related paper, we describe the results of empirical analyses to instantiate the constant parameters of the metareasoning-partition models for the case of belief-network reformulation [4]. Our analyses shall center on solution-planning problems. In solution-planning problems, effort is expended on the meta-analysis of the problem instance before the solution is executed. In related work on the metareasoning-partition problem, Shekhar and Dutta have examined the tradeoff between solution planning and execution for sample search algorithms; these investigators applied their results to enhance the speed of database queries [12]. Research on more general forms of metareasoning, that are interleaved with base-level reasoning, is described in [9, 11]. We shall touch on uncertainty issues, but will remain focused largely on the analyses of deterministic cases; a discussion of resource partition under uncertainty is found in [3].

## 2 Metareasoning Partition Problem

The total time $t$ for generating a solution includes the time used by solution planning and for the execution of the solution. We use $t_m$ to refer to the time used for metareasoning about the problem before executing the solution, and use $t_e$ to refer to the computation time used by base-level execution. The solution of the metareasoning-partition problem is a meta–metanalysis that requires a quantity of reasoning resource itself; the cost of a dynamic analysis of the ideal resource partition is the time required to solve an optimization problem or to apply a previously generated solution. As we shall see, it is feasible to determine the ideal metareasoning resource partition with an inexpensive, constant-cost analysis. We include a constant term $t_{mm}$ in our metareasoning optimization equations to address, in a reflective manner, the cost of solving those equations. Thus, the total time used by the base-level execution, metareasoning, and optimization of the metareasoning partition is

$$t = t_e + t_m + t_{mm}$$

Solution-planning methods are precursory metalevel analyses that increase the efficiency of the execution of a naive base-level reasoning strategy. We can model the relationship between solution planning and execution with functions that capture the efficacy of solution planning for enhancing base-level inference. We shall examine the ideal metareasoning partition for utility-directed and goal-directed problem solving.

### 2.1 Utility-Directed Computation

*Utility-directed* computation centers on the optimization of the value of computation, as dictated by possible actions and an explicit or implicit model of preference. This class of problem solving is receiving increasing attention among AI investigators [7, 9, 2, 11]. With utility-directed computation, we measure and represent the expected value of computation with a numerical measure of preference, termed *utility*. Utility is defined by the axioms of utility theory enumerated by von Neumann and Morgenstern over four decades ago [13].[1]

---

[1] A detailed review of past, and more recent, efforts to apply probability and decision theory for solving challenging artificial-intelligence problems is found in [8].

We use *comprehensive value $u_c$* to refer to the *expected utility* associated with the application of a computational strategy. The comprehensive value can often be decomposed into two components: the *object-related* value and *inference-related* value. The *object-related* value $u_o$ is the expected utility associated with the best action or result available to an agent, given a state of the world. The *inference-related* cost $u_i$ is the expected cost associated with delay of computation. If we can decompose $u_c$ into $u_o$ and $u_i$, and these components of utility are related through addition, the expected value of computation (EVC) is just the difference between the increase in object-level utility and the cost of the additional computation.

A metareasoning-partition analysis of utility-directed computation considers the ability of solution planning to change the rate at which utility is delivered with execution time. Metareasoning partition analyses for utility-directed computation are based on the maximization of the utility of computation with respect to a model solution planning and execution

$$\max_{t_e, t_m} u_c(t_m, t_e) = \max_{t_e, t_m}[u_o(t_m, t_e) - u_i(t_m + t_e + t_{mm})] \qquad (1)$$

We shall introduce and analyze expressive models for the ideal partition of resources for utility-directed computation in Section 3.

## 2.2 Goal-Directed Computation

*Goal-directed* computation refers to a more traditional approach to problem solving, centering on the computation of discrete, predefined results from problem instances. A goal-directed reasoning system seeks to minimize the total time required to compute important goals. Given the functional relationship between $t_e$ and $t_m$, our goal is to identify a $t_m^*$ that minimizes $t$; thus, our metareasoning partition analyses for goal-directed computation focus on the minimization of the total time to solution

$$\min_{t_e, t_m}(t_e + t_m + t_{mm}) \qquad (2)$$

We shall examine and solve models of ideal metareasoning partition for goal-directed systems in Section 4.

# 3 Partition for Utility-Directed Problems

In this section, we will introduce models that can be used to represent and solve the problem of ideally apportioning resource to solution planning for utility-directed problems. We will focus our attention on utility optimization with flexible reasoning strategies, described in Sections 3. We shall first examine the problem of determining ideal problem-solving cessation, in Subsection 3.2. In Subsection 3.3, we will examine the case of ideal partitioning of resources, making use of tools from the ideal cessation problem.

## 3.1 Flexible Computation

A important aspect of developing utility-directed reasoners that are resilient to uncertain challenges and resources is the development of *flexible* or *anytime* computation strategies [5, 6, 2].

We formally analyzed properties of flexible reasoning that are desirable for reasoning under bounded resources [7]. First, we wish our solution strategies to deliver some object-level return for any quantity of allocated resource. Thus, we desire our strategies to provide us with partial results that have value that increases monotonically with allocated resource. We also desire our strategies also to exhibit graceful degradation, or to be relatively insensitive to small reductions in the allocated resource. That is, we wish to limit the magnitude of discontinuous jumps in the value of a partial result as we diminish the quantity of resources devoted to a problem, preferring instead a degree of incrementality or *continuity* in the refinement of partial results with the application of resources. Finally, we wish our results to demonstrate *convergence* on an ideal answer with sufficient resources.

Flexible computation is especially useful for reasoning under uncertain challenges and deadlines; flexible problem-solving generates immediate object-level returns on small quantities of invested computation, and minimizes the risk of dramatic losses in situations of uncertain resource availability. We highlighted these benefits in an analysis of several classes of varying and uncertain cost and deadline [7].

Flexible algorithms for inference have been constructed. These include the bounded-conditioning approach [10] that produce, and iteratively tighten, upper and lower bounds on probabilities of interest for probabilistic inference and decision making under uncertainty.

## 3.2   Ideal Reflection with Flexible Computation

We shall first consider a special case of solution planning that addresses the problem of *ideal reflection*. We wish to determine the ideal time for dwelling on a problem execution before acting. We wish to compute the ideal fraction of the problem to solve, given a flexible reasoning strategy that works to refine an answer as increasing amounts of resource is applied.

Let us consider an example of a medical-diagnosis problem under time constraints, borrowed from [6]. Assume we have a flexible strategy for the generation of a result needed for making a time-pressured medical therapy decision. Figure 2 displays the object-level utility, $u_o$, delivered by a flexible strategy as a result is refined with additional time $t$. To reason about optimal allocation of resources to this strategy, we must also consider the costs of delay. Assume that we find, through preference assessment, that a patient incurs a linear cost when a clinician delays an action in a particular context. Figure 2 shows a specific linear inference-related cost function $u_i(t) = ct$. Assuming that the inference-related cost and the object-level utility are decomposable, and are related by addition, we can determine the comprehensive utility, $u_c$, by adding the object-level utility and the inference-related cost. As portrayed in Figure 2, $u_c$ rises to a maximum, $u_c^*$, at an ideal stopping time $t^*$. After reaching $u_c^*$, it is not worthwhile to expend additional resource: for each tick of the clock past $t^*$, we lose more in the cost of delay than we gain through additional object-level refinement.

We can summarize the relationships among the components of the comprehensive utility, and supply a reasoner with the ability to perform optimizations quickly. For example, if the object-level utility and inference-related costs are related by addition, we know that maxima for $U_c$ will occur when the derivatives

$$\frac{\partial u_o}{\partial t} = -\frac{\partial u_i}{\partial t} \qquad (3)$$

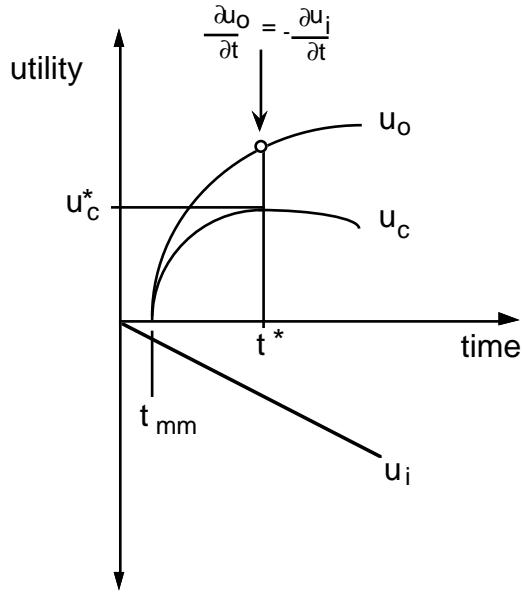

Figure 2: The economics of flexible computation. The graph highlights the fundamental relationships among the object-level value ($u_o$), the cost of reasoning ($u_i$), and the comprehensive value of computation ($u_c$). As indicated by the graph, the ideal halting time ($t^*$) for this utility model is the time where the magnitude of the derivatives of the object-level value and inference-related cost are equal.

If the second derivative of the object-level utility function is everywhere negative and the second derivative of the cost of computation is everywhere nonnegative, the single maximum will indicate the optimal $u_c^*$. Otherwise, we may have to compare several allocations of resource to distinguish local from global maxima.

### 3.2.1 Exponential Model

We now shall develop and solve functions that can model problem solving with flexible computation. Consider the example where we can describe the way the object-level value approaches the value of a complete solution with a negative exponential process

$$u_o(t_e) \;=\; 1 - e^{-kt_e} \tag{4}$$

Such a computational process produces results that are refined incrementally and monotonically, and that converge with some quantity of resource on an ideal object-level answer. Let us assume that the cost of delay is separable from the object-level utility and that the object-level value and cost are related by addition, and that we have a linear cost with

$$u_i(t_e) \;=\; -c(t_e + t_{mm}) \tag{5}$$

where $t_{mm}$ is the constant cost of the optimization process needed to determine the optimal halting point. We know that the comprehensive utility is just the sum of Equations 4 and 5

$$u_c(t_e) \;=\; 1 - e^{-k(t_e)} - c(t_e + t_{mm}) \tag{6}$$

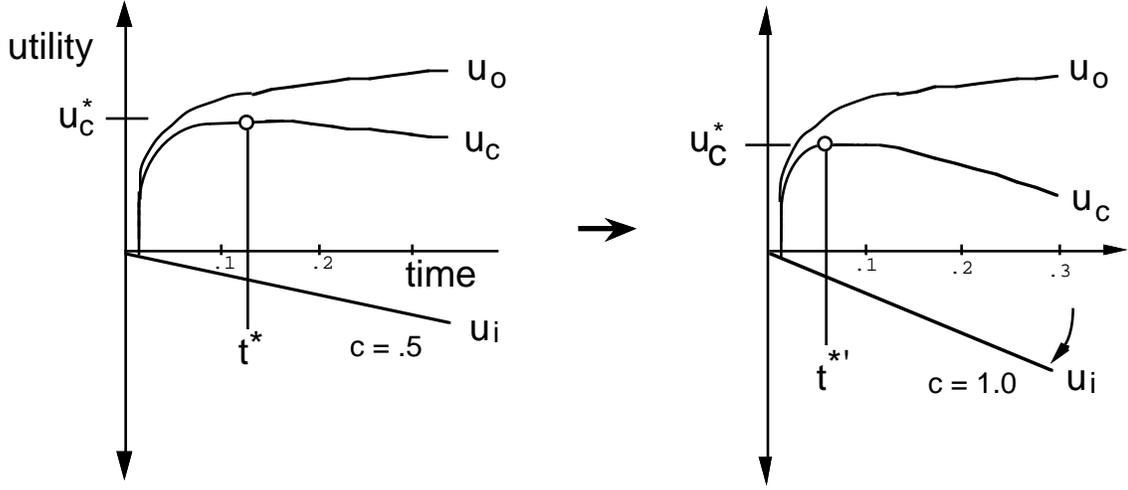

Figure 3: Analysis of the inverse-power model. The graph displays the change in ideal value $u_c^*$ and halting time ($t^*$), given changes in the cost of reasoning ($c$). Here, we change the criticality of the situation by doubling the cost of delay.

We can solve for an optimal halting time by differentiating Equation 6 in terms of the time expended on the problem. For our example, the ideal amount of execution time before halting is

$$t_e^* = -\frac{ln(\frac{c}{k})}{k}$$

and the total time before computation cessation is

$$t^* = -\frac{ln(\frac{c}{k})}{k} + t_{mm}$$

Through substitution of the ideal execution time, and cost of our analysis, into the comprehensive utility function, we can calculate the ideal $u_c$, $u_c^*$. Note that we could substitute the derivative of an arbitrary monotonic cost function, $\mathcal{C}'(T)$, for $c$ in the above analysis, yielding formulae for more complex cost functions.

### 3.2.2 Inverse-Power Model

Flexible computation computation can also be modeled by an inverse-power refinement trajectory, as in the following form

$$u_o(t_e) = 1 - \frac{1}{kt_e^a} \qquad (7)$$

Assuming a linear cost $c$ with time, the ideal amount of execution time before halting for this model is

$$t_e^* = \left(\frac{a}{kc}\right)^{\frac{1}{a+1}}$$

and $u_c^*(t_e)$, at the ideal halting point, $t^*$, is

$$u_c^*(t_e) = 1 - k^{-1}\left(\frac{a}{kc}\right)^{-\frac{a}{a+1}} - c(t_e^* + t_{mm}) \qquad (8)$$

With a minimal expenditure of resources, we can quickly determine the ideal computation time and utility, as a function of a small number of problem-solving parameters.

The graph at the left of Figure 3 demonstrates how we can use these models to efficiently determine how the optimal halting time and ideal comprehensive utility will change, given changes in the cost of reasoning (dictated by $c$ or, more generally, by a function $\mathcal{C}$), in the rate of refinement (dictated by $k$), and the cost of determining the optimal halting time ($t_{mm}$). In the figure, we change the criticality of the situation by increasing the cost of delay. The new optimization shows that we should reflect for less time, and can expect to receive less value from reasoning.

We have studied the behavior of several prototypical equations. As an example, consider the case of problem-solving described by Equation 6, where $k = .1$, and the metareasoning time required for the optimization of execution dwell, $t_{mm}$, is .01 seconds. In this situation, when the urgency, represented by the cost $c$, is .04, the ideal execution time, $t_e^*$, is 9.16 seconds, and the optimal comprehensive utility of computation, $u_e^*$, is .23. If we double the cost of delay, so that $c = .08$, we find that $t_e^*$ is reduced to 2.23 seconds and $u_c^*$ drops to .02.

## 3.3   Ideal Metareasoning Partition

In the previous analysis, we assumed the absence of a capability for solution planning to make execution more efficient. We studied a solution planner that could only make decisions about when to halt base-level computation. Now, let us enrich the analyses described in Section 3.2 to represent the ability of a metareasoner to enhance the efficacy of object-level problem solving. We can model metareasoning processes with functions $\mathcal{K}(t_m)$ that modify the rates of problem solving in terms of the resources applied to meta-analysis. In particular, let us model the efficacy of solution planning with functions that dictate the solution-refinement constants $k$ in our object-level utility models as monotonic functions of the amount of time expended for metareasoning. That is, we make $k$ a variable that is a function of the planning time where

$$\mathcal{K}(t_m) \;=\; k$$

In cases where we are uncertain about the efficacy of solution planning, our analysis can be extended to consider a probability distribution over $k$, $p(k|t_m)$. For example, an enriched model for the exponential form, defined in Equation 6, is

$$u_o(t_m, t_e) \;=\; 1 - e^{-\mathcal{K}(t_m)t_e} \tag{9}$$

Now, the rate of refinement depends on the amount of solution planning. Similarly, for the inverse-power form, we add a $\mathcal{K}(t_m)$ that is sensitive to solution planning,

$$u_o(t_m, t_e) \;=\; 1 - \frac{1}{[\mathcal{K}(t_m)]^b t_e^a} \tag{10}$$

Let us delve more deeply into the inverse-power model of solution planning. Our goal now is to maximize the following model of comprehensive utility with respect to solution planning and execution

$$u_c(t_m, t_e) \;=\; 1 - \frac{1}{[\mathcal{K}(t_m)]^b t_e^a} - \mathcal{C}(t_m + t_e + t_{mm}) \tag{11}$$

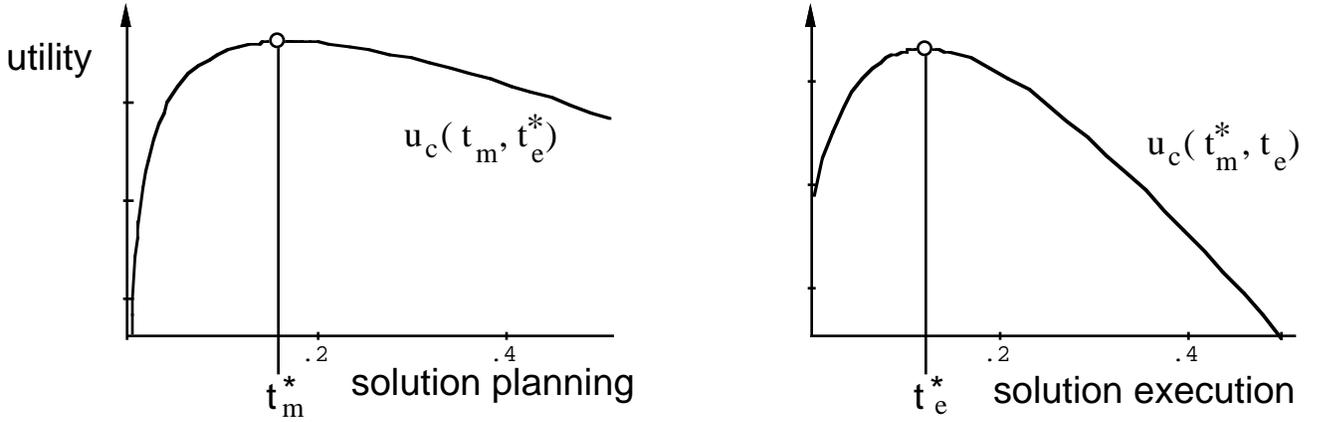

Figure 4: The structure of the optimum. Graph (a) displays the sensitivity of the utility of computation to variation of the time for metareasoning, given an ideal time for execution of the solution. Graph (b) displays the sensitivity of the utility of computation to variation of the time for execution, given an optimal partition of resources for solution planning.

To calculate the ideal amount of time $t_m$ to apply to the process of solution planning, we differentiate Equation 11, in terms of $t_e$ and $t_m$, and solve the equations simultaneously. We identify the constraint that

$$t_e^* = \frac{a}{b} \frac{\mathcal{K}(t_m)}{\mathcal{K}'(t_m)} \tag{12}$$

Through resubstituting this result, we obtain

$$\mathcal{K}(t_m) = \left( \frac{[b(\mathcal{K}'(t_m))]^{a+1}}{a^a C'(t_m)} \right)^{\frac{1}{a+b+1}} \tag{13}$$

We can substitute different $t_m$ into Equation 13 and solve for $t_m^*$.[2] We can then determine the optimal execution time, $t_e^*$, and the ideal value of computation, $u_c^*$, in terms of $t_m^*$. Closed-form solutions to Equation 13 can be identified for many families of $\mathcal{K}(t_m)$. Let us derive the ideal planning time and ideal value of computation for the case where the efficacy of solution planning is modeled as a linear increase in $k$ with planning time

$$\mathcal{K}(t_m) = k_o + l t_m$$

where $l$ is a metareasoning-efficiency constant. If we assume a linear cost $c$ with total time to solution, it follows from Equation 13 that

$$t_m^* = l^{-1} \left[ \left( \frac{(bl)^{a+1}}{ca^a} \right)^{\frac{1}{a+b+1}} - k_o \right] \tag{14}$$

and

$$t_e^* = \frac{a}{b} \left( \frac{k_o}{l} + t_m^* \right) \tag{15}$$

___
[2]The existence of a non-zero maxima can be confirmed by checking that the Hessian matrix for the utility function is negative semidefinite at the solution points.

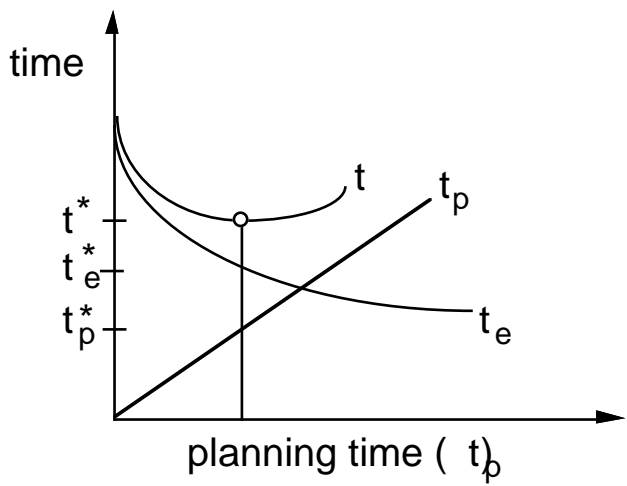

Figure 5: A graph demonstrating relationships between the total time required for solving a goal-directed problem ($t^*$), and time allocated for solution planning ($t_m^*$) and execution ($t_e^*$).

Thus, the ideal planning time and ideal execution time can be quickly calculated from a small number of constant parameters of our model and a constant dictating the urgency of a situation. The ideal value, reached with a system with capabilities described by this model, follows from substituting the ideal solution-planning and execution times into Equation 11.

# 4 Partition for Goal-Directed Problems

Let us now move from the case of optimizing the value of flexible computation to the special case of minimizing the amount of time required to generate a predefined result. We represent the efficacy of solution planning with a function $\mathcal{P}$ that reports the expected amount of time required for solving a problem, as a function of of the amount of time spent planning the solution strategy. Thus,

$$\mathcal{P}(t_m) = t_e \tag{16}$$

In the general case, we may be uncertain about the relationships among $t_m$ and $t_e$, and, thus, uncertain about $t$. In these situations, we are interested in assessing and applying knowledge in the form of probability distributions,

$$\mathcal{P}(t_m) = p(t_e|t_m) \tag{17}$$

conditioned on different amounts of solution planning. We describe solutions to goal-directed computation under uncertainty in [3]. For the deterministic case, the total time for reasoning needed before applying a particular result in the world is

$$t = \mathcal{P}(t_m) + t_m + t_{mm} \tag{18}$$

We seek to minimize $t$. To determine the optimal partition, we differentiate Equation 18 with respect to solution-planning time and search for a minimum. Thus, we explore solution-planning times where

$$\frac{\partial \mathcal{P}(t_m)}{\partial t_m} = -1 \tag{19}$$

The problem-reduction function $\mathcal{P}(t_m)$ can take on different forms, depending on the relationship identified between execution efficiency and the extent of metareasoning analysis. For example, for the case where solution planning can be expected reduce the complexity of the problem through reformulation, $\mathcal{P}(t_m)$ dictates the reduction in complexity of generating a complete solution as a function of planning time. A problem-reduction function also can represent the expected increases in the rate at which a flexible-computation policy refines a problem. We shall examine this situation for a simple example.

Assume that the abilities of our base-level problem-solver are described by an exponential model described in Equation 9. Unlike our utility-directed examples, where we sought to optimize utility, our goal now is to generate a predefined goal. Assume that our goal is to compute until a result, worth some fraction $f$ of the complete value of a problem, is obtained. We are interested in minimizing the total time of delay, given

$$1 - e^{-\mathcal{K}(t_m)t_e} = f \tag{20}$$

Thus, we wish to minimize Equation 18, subject to the constraints of Equation 20.

$$\mathcal{P}(t_m) = t_e = -\frac{ln(1-f)}{\mathcal{K}(t_m)} \tag{21}$$

Differentiating 21 with respect to solution-planning time, and solving for minima, we find that $t_m^*$ satisfies the constraint

$$\frac{(\mathcal{K}[t_m])^2}{\mathcal{K}'(t_m)} = -ln(1-f) \tag{22}$$

We can verify that we are at a minimum by confirming that the second derivative of the equation for total time is positive at the identified $t_m$. For a model where $\mathcal{K}(t_m) = K_o + lt_m$, the ideal solution-planning time is

$$t_m^* = \frac{[k^2 - l(k^2 + ln[1-f])]^{\frac{1}{2}} - k}{l} \tag{23}$$

This function tells us that the ideal portion of resources to apply to metareasoning is a simple function of the solution-refinement constant and the desired quality of the goal.

# 5   The Value of Metareasoning

We can compare the relative optimality of agents of different constitutions and capabilities, as well as the effectiveness of different policies for reasoning and metareasoning. As an example, we can determine the value of adding a solution-planning capability to systems that ideally apply flexible reasoning strategies. For a single case, the value of adding a solution-planning capability for agents, modeled by the inverse-power models presented in Sections 3.2 and 3.3, is just the difference between the ideal value available with and without the additional planning capabilities. We can generalize our analysis, from comparisons of policies and capabilities for specific instances, to the value of an agent over time. We assume that an agent is immersed in an environment for some length of time (e.g., the expected lifetime of the agent).

In considering a set of challenges, it can be essential to consider probability distributions that describe the difficulty of problems and the cost associated with different challenges. Let us assume an *independent-challenge model*, asserting that solutions to individual challenges do not significantly interact [6]. Assume that $\mathcal{A}_1$ is an agent with ideal reflection and solution planning, $\mathcal{A}_2$ is an agent with an ideal-reflection policy, and $\mathcal{I}_i$ is the problem instance at hand. Each problem $\mathcal{I}_i$ is associated with a tuple $(c_i, k_i)$. $k_i$ is an expected base problem-solution rate ($k_o$) for problem $i$. Harder problems are associated with smaller $k_i$ and, thus, are solved more slowly. $c_i$ is a linear cost constant (or cost function) incurred with delay in addressing problem $\mathcal{I}_i$. The utility gains of adding a planning capability to our agent with an ideal reflection capability, can be captured by substituting these constants into the solution of the ideal comprehensive utilities the analyses in Sections 3.2 and 3.3.[3] For an environment described by a probability distribution, $p(i)$, and where the frequency that problems challenge an agent is $\mathcal{F}$,

$$\Delta v_a[p(\mathcal{I}_i), \mathcal{F}, t] \;\; = \;\; t\mathcal{F} \int \left[ u_o^*(\mathcal{A}_1, \mathcal{I}_i) - u_o^*(\mathcal{A}_2, \mathcal{I}_i) \right] p(\mathcal{I}_i) di$$

# 6    Summary and Conclusions

We have constructed and analyzed models that capture fundamental relationships between metareasoning and base-level problem solving. In particular, we exercised general functional forms, relating metareasoning and execution efficiency for utility-directed and goal-directed computation. We identified solutions to metareasoning-partition problems as functions of a small number of constant parameters of the models. The models and solutions for the ideal apportionment of resources to solution planning and execution suggest how knowledge about the efficacy of solution planning can be used to endow agents with an ability to efficiently custom-tailor their use of resources. Finally, we discussed the value of metareasoning in an environment characterized by a distribution over problems, each associated with a difficulty and a cost. Empirical work is underway, within the realm of belief-network reformulation, to acquire knowledge about the efficacy of solution planning, and to represent that knowledge with specific parameterizations of our resource-partition models.

# Acknowledgments

We thank Greg Cooper, Tom Dean, and Mathew Ginsberg for stimulating discussions on metareasoning.

# References

[1] J.A. Barnett. How much is control knowledge worth? *Journal of Artificial Intelligence*, 22(1):77–89, January 1984.

---

[3]For simplicity, we do not include time-discounting of future gains and losses here; time discounting can be captured by including a factor that discounts future gains as a function of time.

[2] M. Boddy and T. Dean. Solving time-dependent planning problems. In *Proceedings of the Eleventh IJCAI*. AAAI/International Joint Conferences on Artificial Intelligence, August 1989.

[3] J.S. Breese and E.J. Horvitz. Principles of problem reformulation under uncertainty. Technical report, Knowledge Systems Laboratory, Stanford University, February 1990. KSL-90-27.

[4] J.S. Breese and E.J. Horvitz. Reformulating and solving belief networks under bounded resources. Technical report, Knowledge Systems Laboratory, Stanford University, March 1990. KSL-90-28.

[5] T. Dean and M. Boddy. An analysis of time-dependent planning. In *Proceedings AAAI-88 Seventh National Conference on Artificial Intelligence*, pages 49–54. American Association for Artificial Intelligence, August 1988.

[6] E.J. Horvitz. Reasoning about beliefs and actions under computational resource constraints. In *Proceedings of Third Workshop on Uncertainty in Artificial Intelligence*, Seattle, Washington, July 1987. American Association for Artificial Intelligence. Also in L. Kanal, T. Levitt, and J. Lemmer, ed., *Uncertainty in Artificial Intelligence 3,* Elsevier, 1989, pps. 301-324.

[7] E.J. Horvitz. Reasoning under varying and uncertain resource constraints. In *Proceedings AAAI-88 Seventh National Conference on Artificial Intelligence*, pages 111–116. American Association for Artificial Intelligence, August 1988.

[8] E.J. Horvitz, J.S. Breese, and M. Henrion. Decision theory in expert systems and artificial intelligence. *International Journal of Approximate Reasoning*, 2:247–302, 1988. Special Issue on Uncertain Reasoning.

[9] E.J. Horvitz, G.F. Cooper, and D.E. Heckerman. Reflection and action under scarce resources: Theoretical principles and empirical study. In *Proceedings of the Eleventh IJCAI*, pages 1121–1127. AAAI/International Joint Conferences on Artificial Intelligence, August 1989.

[10] E.J. Horvitz, H.J. Suermondt, and G.F. Cooper. Bounded conditioning: Flexible inference for decisions under scarce resources. In *Proceedings of Fifth Workshop on Uncertainty in Artificial Intelligence*, Windsor, Canada, August 1989. American Association for Artificial Intelligence.

[11] S.J. Russell and E.H. Wefald. Principles of metareasoning. In Ronald J. Brachman, Hector J. Levesque, and Raymond Reiter, editors, *Proceedings of the First International Conference on Principles of Knowledge Representation and Reasoning*, Toronto, May 1989. Morgan Kaufman.

[12] S. Shekhar and S. Dutta. Minimizing response times in real time planning and search. In *Proceedings of the Eleventh IJCAI*, pages 238–242. AAAI/International Joint Conferences on Artificial Intelligence, August 1989.

[13] J. von Neumann and O. Morgenstern. *Theory of Games and Economic Behavior*. Princeton University Press, Princeton, NJ, 1947.